\pdfoutput=1
\documentclass{article} %

\usepackage{nips14submit_e,times}
\usepackage{acronym}
\usepackage{todonotes}
\usepackage{graphicx}
\usepackage{siunitx}
\usepackage{amsmath}
\usepackage{caption}
\usepackage{subcaption}
\usepackage{hyperref}
\usepackage{color,soul}
\pdfoutput=1

\newcommand{\eg}{\textit{e.g. }}
\newcommand{\ie}{\textit{i.e. }}
\newcommand{\comment}[1]{}

\newcommand{\BNETE}[0]{{\bf RetiNet~B (extreme)}}
\newcommand{\BNET}[0]{{\bf RetiNet~B}}
\newcommand{\RTNET}[0]{{\bf RetiNet~C}}
\newcommand{\VGG}[0]{{\bf VGG19}}
\newcommand{\DSNET}[0]{{\bf DenseNet}}
\newcommand{\RESNET}[0]{{\bf ResNet}}
\newcommand{\VNET}[0]{{\bf 2DSeg}}

\providecommand{\keywords}[1]{\textbf{\textit{keywords ---}} #1}
\nipsfinalcopy 

\begin{document}

\title{RetiNet: Automatic AMD identification \\in OCT volumetric data}

\author{S.~Apostolopoulos, C.~Ciller, S.~De~Zanet, S.~Wolf~and~R.~Sznitman	\thanks{S.~Apostolopoulos and R.~Sznitman are with  the ARTORG Center, University of Bern, Switzerland. E-mail: firstname.lastname@artorg.unibe.ch}
	\thanks{C.~Ciller is with the Radiology Department, CIBM, Lausanne University and University Hospital, Lausanne and with the Ophthalmic Technology Group, ARTORG Center Univ. of Bern, Switzerland}
	\thanks{S.~De~Zanet is with the Ecole Polytechnique Federale de Lausanne, Switzerland.}
	\thanks{S.~Wolf is with the Bern University Hospital, Inselspital, Switzerland.}
}

\maketitle

\begin{abstract}
Optical Coherence Tomography (OCT) provides a unique ability to image the eye retina in 3D at micrometer resolution and gives ophthalmologist the ability to visualize retinal diseases such as Age-Related Macular Degeneration (AMD). While visual inspection of OCT volumes remains the main method for AMD identification, doing so is time consuming as each cross-section within the volume must be inspected individually by the clinician. In much the same way, acquiring ground truth information for each cross-section is expensive and time consuming. This fact heavily limits the ability to acquire large amounts of groundtruth, which subsequently impacts the performance of learning-based methods geared at automatic pathology identification. To avoid this burden, we propose a novel strategy for automatic analysis of OCT volumes where only volume labels are needed. That is, we train a classifier in a semi-supervised manner to conduct this task. Our approach uses a novel Convolutional Neural Network (CNN) architecture, that only needs volume-level labels to be trained to automatically asses whether an OCT volume is healthy or contains AMD. Our architecture involves first learning a cross-section pathology classifier using pseudo-labels that could be corrupted and then leverage these towards a more accurate volume-level classification. We then show that our approach provides excellent performances on a publicly available dataset and outperforms a number of existing automatic techniques. 
\end{abstract}


\maketitle

\acrodef{BM}{Bruch\'{}s Membrane}
\acrodef{ILM}{Inner Limiting Membrane}
\acrodef{OCT}{Optical Coherence Tomography}
\acrodef{DoG}{Difference of Gaussians}
\acrodef{RANSAC}{RANdom SAmpling Consensus}
\acrodef{ReLU}{Rectified Linear Unit}
\acrodef{AMD}{Age-Related Macular Degeneration}
\acrodef{DR}{Diabetic Retinopathy}
\acrodef{DME}{Diabetic Macular Edema}
\acrodef{RPE}{Retinal Pigment Epithelium}
\acrodef{CNN}{Convolutional Neural Network}
\acrodef{MP}{Max Pooling}
\acrodef{AP}{Average Pooling}
\acrodef{BN}{Batch Normalization}
\acrodef{BSL2D}{Baseline 2D}
\acrodef{SD-OCT}{Spectral-Domain Optical Coherence Tomography}
\acrodef{Fundus}{Fundus Image Photography}
\acrodef{ROI}{Region of Interest}
\acrodef{HoG}{Histogram of Gradients}
\acrodef{BoW}{Bag of Words}
\acrodef{SVM}{Support Vector Machine}
\acrodef{LBP}{Local Binary Patterns}
\acrodef{PCA}{Principal Component Analysis}
\acrodef{IRC}{Intra-retinal Cysts}
\acrodef{SRF}{Subretinal Fluid}
\acrodef{AUC}{Area Under the Curve}
\acrodef{ROC}{Receiver Operating Characteristic}
\acrodef{BM3D}{Block Matching and 3D filtering}

\section{Introduction}

By and large, Optical Coherence Tomography (OCT) has reshaped the field of ophthalmology ever since its inception in the early 90s~\cite{Huang1991}. At its core, OCT uses infrared-light interferometry to image through tissue in order to characterize anatomical structures beyond their surface. Given its simplicity, affordability and safety, it is no surprise that its use has gained widespread popularity for both disease diagnosis and treatment. Similarly, its use has gained traction in other medical fields such as for histopathology and skin cancer analysis~\cite{Welzel2001}. 

\begin{figure}
\centering
\includegraphics[width=0.8\textwidth]{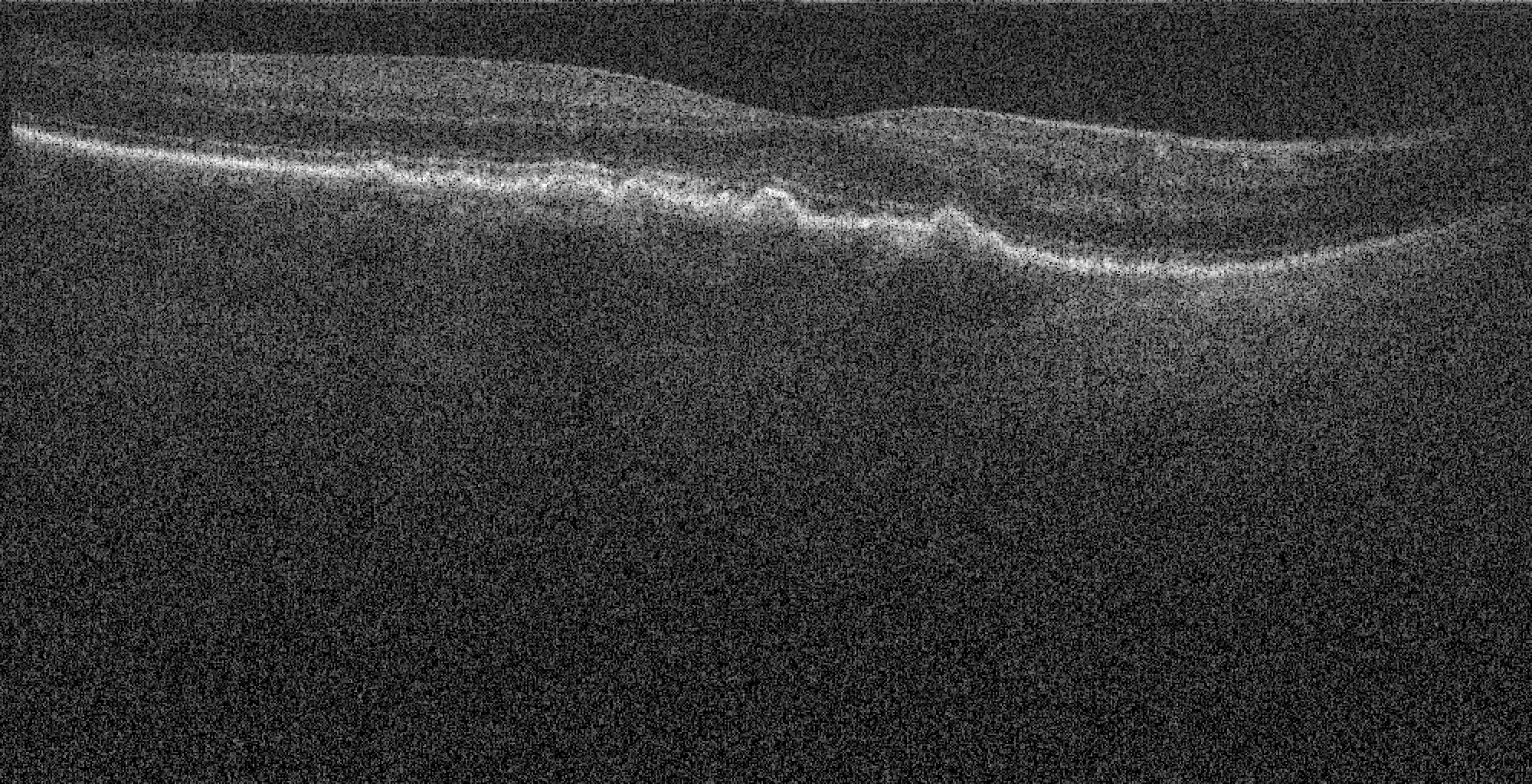}
\caption{An example of a B-scan cross-section of a patient with \ac{AMD} in the foveal pit area. Visible are the multiple retinal layers, including the Retinal Pigment Epithelium (RPE) and Bruch\'{}s Membrane (BM). The latter is perturbed with \textit{drusen}, which manifest as bumps disrupting this continuous layer.}
\label{ref:oct_example}
\end{figure}

Indeed, with an ability to image the posterior part of the eye in 3D (\eg the retina) at micrometer resolution, OCT imaging now allows for visualization of most retinal layers~\cite{Abramoff2010,Garvin2009} and more importantly, numerous pathological markers, such as intraretinal fluid, drusens or cysts~\cite{Farsiu2014,Jager2008}. As illustrated in Fig.~\ref{ref:oct_example}, such markers can be observed in OCT cross-sectional images, or {\it B-scans} and have been linked to a number of eye conditions, including \ac{AMD} and Diabetic Retinopathy (DR) which currently affect over 8.7\% of the world population and 159 million people worldwide, respectively~\cite{Farsiu2014,Yau2012,Wong2014}. Moreover, these pathologies are the major cause of blindness in developed countries~\cite{Bressler2004}. Alarmingly, the number of people with either of these diseases is projected to skyrocket, with \ac{AMD} affecting an estimated 196 million people by 2020 and 288 million people by 2040~\cite{Wong2014}. Genetic factors, race, smoking habits and the ever growing world population are responsible for this pathology growth~\cite{AMDResearchGroup2000}.

While OCT has gained significant importance in recent years for AMD and DR screening~\cite{Bressler2002,Ciulla2003}, the process to do so remains time consuming however. In effect, 3D OCT volumes, also referred to as {\it C-scans}, are comprised of 50-100 cross-sectional B-scans. Traditionally, inspection of each B-scan is necessary in order to properly rule-out most retinal diseases. This process is particularly tedious not only due to its time-consuming nature, but also due to the multiple cross-sections that need to be inspected simultaneously to identify elusive and scarce traces of early-stage ocular diseases. In this context, automated algorithms for pathology identification in OCT volumes would be of great benefit for clinicians and ophthalmologists, as access to OCT devices becomes common and nation-wide screening programs commence~\cite{SchmidtErfurth2014}.

Recently, research has given way to a variety of image processing methods for OCT imaging. Some of these have included: techniques for image denoising~\cite{Adler2004,Dabov2006,Maggioni2013}, strategies for improved image reconstruction~\cite{Wojtkowski2004,Zawadzki2007,Kraus2012,Szkulmowski2013,Montuoro2014,Steiner2015}, dosimetry laser control systems~\cite{Muller2012,Steiner2016,Serife2016} or instrument detection during surgical procedures~\cite{Tao2014,El-haddad2015}.

More specific to pathology identification, various groups have explored automatic detection of retinal pathologies using machine learning techniques, either focusing on segmentation of relevant pathological markers~\cite{Quellec2010,Chiu2012,Dufour2013,Venhuizen2015b,Schlegl2015} or classification of 2D B-scans or 3D C-scans~\cite{Liu2010,Srinivasan2014,Schlegl2015,Venhuizen2015,Lemaitre2016}. While effective to some extent, most of these works have leveraged B-scan level groundtruth information in order to learn classification functions. These more detailed labels are unfortunately often not available and as such, limit the usability of these solutions. 

To this end, we present a new strategy towards automatic pathology identification in OCT C-scans using only volume level annotations. To do this, we introduce a novel Convolution Neural Network (CNN) architecture, named {\it RetiNet}, that directly estimates the state of a C-scan solely using the image data and without needing additional information. At its core, our approach uses (1) a task-specific volume pre-processing strategy where we flatten and normalize the data in an OCT-specific manner, (2) we then train a 2D B-scan CNN using pseudo-labels that could be {\it corrupted} in order to pre-learn filters that respond to relevant image features and (3) reuse the learned features in a C-scan level CNN that takes a mosaic of B-scans as input and classifies the entire C-scan at once. Using a publicly available OCT dataset~\cite{Farsiu2014}, we show that our approach is highly effective at separating AMD from control subjects and outperforms existing state-of-the-art methods for image classification. In addition, we not only show that RetiNet outperforms excellent recent networks from the computer vision literature trained from scratch, but also surpasses the performance of state-of-the-art pre-trained networks with adapted filters. Last, we show how our approach provides high performances in terms accuracy, learning pathology-specific filters capable to identifying pathological markers effectively. 	

The remainder of this article is organized as follows: The following section discusses the relevant related work. Sec.~\ref{sec:method} then describes in detail our approach and the RetiNet architecture. Following this, we describe our experimental section and the evaluation of several baseline strategies in Sec.~\ref{sec:experiments}. We then conclude with final remarks in Sec.~\ref{sec:conc}.

\section{Related Work}
\label{sec:related}

We now briefly discuss a number of related works on the topic of OCT data classification.

In Venhuizen et al.~\cite{Venhuizen2015} regions of interest are automatically extracted around the center of each C-scan via an intensity threshold. \ac{PCA} is then applied to each region for dimensionality reduction, followed by K-means clustering in order to build a \ac{BoW} representation, which is then used in combination with a Random Forest classifier. The classifier is trained on a set of 284 AMD patients and healthy controls and evaluated on a balanced set of 50 AMD patients and 50 healthy controls. 

The same dataset was previously used by Farsiu et al.~\cite{Farsiu2014}, who developed a semi-automatic classification method for \ac{AMD} patients. Given manually-corrected segmentations of~\ac{BM}, the~\ac{RPE} and ~\ac{ILM} layers, they calculated a number of metrics: total thickness of the retina; thickness between drusen apexes and the \ac{RPE}; abnormal thickness score; abnormal thinness score. From these, they trained linear regression models using different combinations of these metrics.

Srinivasan et al.~\cite{Srinivasan2014} presented a method for classifying \ac{AMD}, \ac{DME} and healthy C-scans using multiscale \ac{HoG} features and a \ac{SVM} classifier. Each B-scan was first resized to a resolution of 246x256 pixels, denoised using the \ac{BM3D} algorithm~\cite{Dabov2006} and then flattened. 
While the dataset for this method is public, it is only available in a preprocessed form, which unfortunately limits our ability to compare to it.

More recently, Lemaitre et al~\cite{Lemaitre2016} followed in the direction of Liu et al.~\cite{Liu2010}, by extracting 2D and 3D \ac{LBP} features from a set of 16 healthy and 16 patients suffering from \ac{DME}. B-scans were denoised using the non-local means (NLM) algorithm~\cite{Buades2005} and flattened. The use of different linear and non-linear classifiers was then explored to identify which performed best. 

Finally, Schlegl et al.~\cite{Schlegl2015} employed a 2D patch-based \ac{CNN} to classify retinal tissue into \ac{IRC}, \ac{SRF} and healthy categories, while providing information to the location of the pathology. They train their classifier using three different ground truths: \textit{weak-labeling}, wherein a single label is applied to the whole C-scan; weak-labeling with \textit{semantic information}, wherein coarse information about the location of the pathology is applied along with the weak label; \textit{full-labeling}, wherein the classifier is trained on the per-voxel ground truth of the whole C-scan. The latter approach yields the best results, with 97.7\%, 89.61\% and 91.98\% for the healthy, \ac{IRC} and \ac{SRF} classes, respectively. While the weak-labeling approach performs significantly worse, at 65.63\%, 21.94\% and 90.30\%, respectively, this setting is the most closely related to the one in the present work.

More specifically, we present a novel method to automatically evaluate~\ac{AMD} or healthy volumes. Our strategy has two important advantages over existing methods: (1) it relies only on volume level labels to be trained and (2) it evaluates complete volumes in one shot, making it simpler to use. As we will show in Sec.~\ref{sec:experiments}, our approach allows for significant performance gains over these existing methods.

\begin{figure*}
\centering
\includegraphics[width=0.99\textwidth]{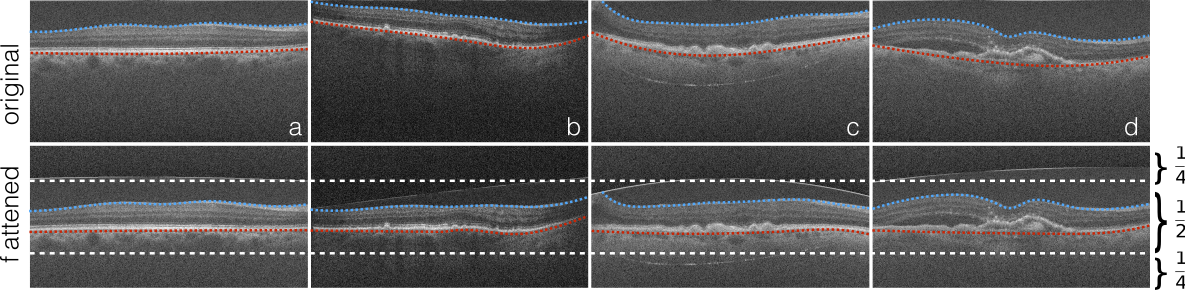}
\caption{Different retina \ac{OCT} B-scans with variations in position and tilt. ({\it{top row}}) Original B-scans with highlighted \ac{ILM} and \ac{BM} layers in blue and red, respectively. ({\it{bottom row}}) Corresponding B-scans after flattening applied. White dotted lines illustrate the regions that are cropped to reduce B-scan sizes. }
\label{fig:flattening}
\end{figure*}

\section{Our Approach}
\label{sec:method}

The overall goal in this work is to automatically evaluate whether an \ac{OCT} volume contains \ac{AMD}. The main challenges in tackling this problem lies in the fact that (1) relatively few volumes are typically available for training classification models even though volumes are large in size (\eg $500 \times 1000 \times 100$ pixels) and (2) that labels denoting the presence of pathology are only available at the volume level and not at the cross-section level.

To perform effective volume classification, we will follow a Deep CNN approach and will describe in the following section our novel architecture to do so. In general, our approach relies on a three-stage process. 

The first is an \ac{OCT}-specific normalization and data-augmentation strategy for \ac{OCT} volumes in order to improve overall generalization and classification performance. Here, we reduce image dimensionality and flatten \ac{OCT} scans in order to regularize the data. Similarly, we make use of symmetries particular to the eyes in order to augment the data effectively. The second stage attempts to learn pathology-specific features at cross-section B-scan levels using volume-level labels. Here, we make use of a relatively simple network to learn filters that are relevant for 2D \ac{OCT} image data. In the last stage, we remap the volume to a large image mosaic and train a new volume-level network by leveraging the previously learned filters that operate at the B-scan level. 

We now begin by formalizing our problem and establish the necessary notation to precisely describe our strategy.

\begin{figure*}[t!]
  \centering
  \includegraphics[width=.99\textwidth]{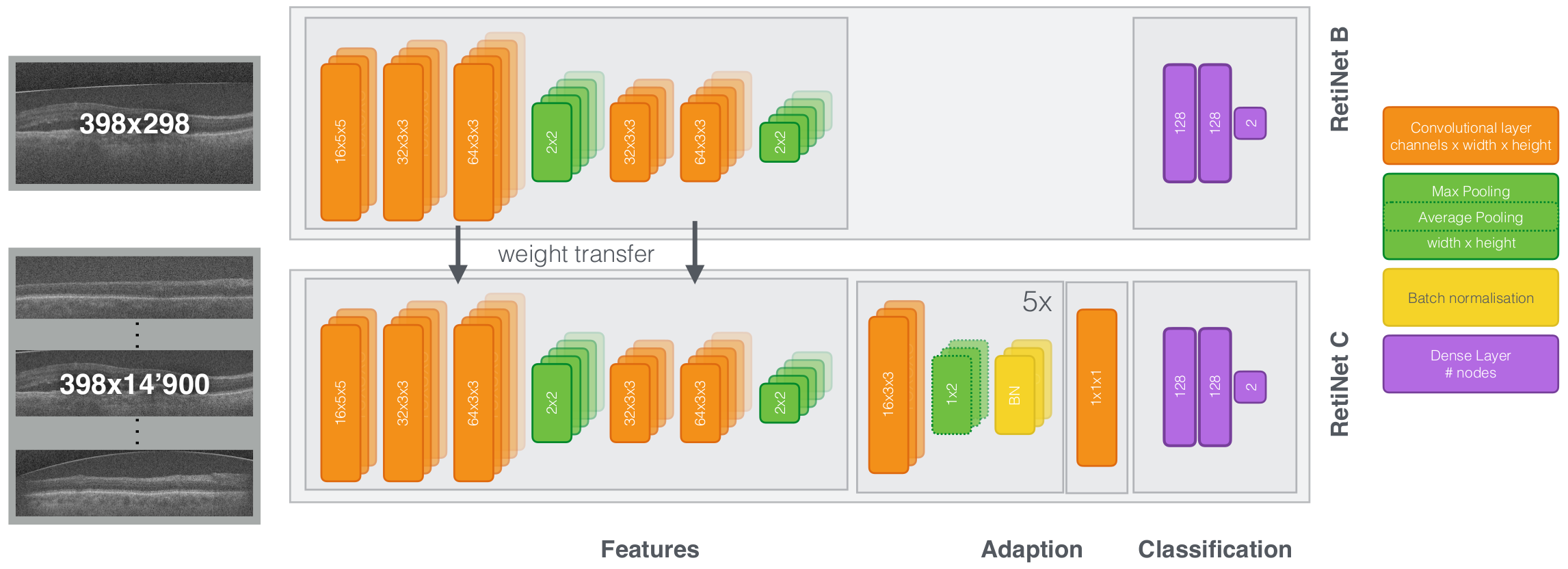}
  \caption{(\textit{top}) \BNET~network layout. Single B-scans are taken as input through the \textit{$FEATURE$} layers and classified in the \textit{$CLASSIFICATION$} layer. Weak labels derived from the the volume label are used to train this network. (\textit{bottom}) \RTNET~network layout. B-scans from an entire \ac{OCT} volume are concatenated vertically into a single image as input. The \textit{$FEATURE$} layers are transferred from \BNET~with their weights to this network. An additional adaption block is added, composed of a repeated layers of convolutions, average pooling and batch normalization. 
  }
  \label{fig:retinet}
\end{figure*}


\subsection{Notation and formulation}
\label{sec:formulation}

Without loss of generality, we assume that our training data $V = \{V_1,\ldots,V_N\}$ is comprised of $N$ OCT volumes. Each volume $V_n$ is of dimension $\{W \times H \times D \} = \mathcal{V}$, where a B-scan cross-section consists of a $W \times D$ image, with $D$ being the depth of the penetrating OCT light source. For a volume $V_n$, we denote $B_n^h, h=1,\ldots,H$, as the $h^{th}$ B-scan in the volume.

Each volume in $V_n$ is associated with a class label $Y_n \in \{0,1\} = \mathcal{Y}$ such that 0 corresponds to control volumes and $1$ corresponds to pathological volumes (\ie \ac{AMD}). Our goal is to learn a classification function $f: \mathcal{V} \rightarrow \mathcal{Y}$ using the training set and labels available. Importantly, we assume that no information is available on the labels of $B_n^h$, as these are expensive to gather.


\subsection{Data preprocessing}
\label{sec:preprocessing}

\subsubsection{Normalization}
\label{sec:normalization}
As can be seen in Fig.~\ref{fig:flattening}~(\textit{top row}), there is both high variability in positions of $B_n^h$ with respect to the anatomy and distortion of the retinal layers. In particular, the retinal layers may be tilted, shifted vertically, distorted due to the acquisition process and of varying intensity. In addition, significant portions of $B_n^h$ images contain little informative content, such as the area above the \ac{ILM} and below the \ac{BM} layer. This is due to the \ac{OCT} imaging device and consists of either noise or regions too deep for the OCT laser to penetrate. 

As such, in order to provide a more compact and consistent set of training volumes, we are interested in normalizing and reducing the size of the \ac{OCT} volumes. Unfortunately, simply cropping each B-scan would be ill-suited, since the retina is curved and this would either remove most of the informative data or result in marginal resizing. To this end, we propose an effective normalization or {\it flattening} strategy.

Our flattening approach consists in aligning the individual BM layers, rectifying for the eye curvature and normalizing for variations in volume intensities. To do so, we first detect the \ac{BM} layer by applying an anisotropic filter on $B_n^h$ using 200 diffusion iterations~\cite{Perona1990}
$$
 c\left(\| \nabla I\| \right) = \frac{1}{1 + \left(\frac{\|\nabla I\|}{\kappa}\right)^2}, 
$$
where $\kappa=50$ was empirically set for all experiments. We then compute the \ac{DoG} from the filtered responses and estimate the \ac{BM} layer as maximal gradient pixels. Naturally, these responses are noisy and incorrect in some cases. For this reason, we fit a second-order polynomial model to the noisy responses using RANSAC outlier detection~\cite{Foley1981}. We then warp the estimated~\ac{BM} to a vertical line centered at 60\% of the image height. In order to reduce the dimension of the image, we resize each $B_n^h$ to be of smaller size $w \times d$, $w<W,d<D$. As can be seen in Fig.~\ref{fig:flattening}, areas above the \ac{ILM} and below the \ac{BM} only contain noise. For this reason, we crop every $B_n^h$ by discarding every voxel $v$ in the C-scan so that $\frac{d}{4} \leq v_d \leq \frac{3d}{4}$, with $v_d$ being the depth of the voxel.

Finally, intensity variations in \ac{OCT} are common when looking at acquisitions over different patients. In order to regularize across these variations, the voxel intensities are normalized to be zero mean and with a standard deviation of one. 

\subsubsection{Data augmentation}
\label{sec:augmentation}

We use data augmentation to increase the number of samples in our training dataset and reduce overfitting~\cite{Simard2003,Krizhevsky2012}. In particular, we take advantage of the bilateral symmetry of the eye to effectively double the number of samples. The resulting samples are biologically plausible, \ie the optic disc, fovea and vessels remain at the correct spots relative to each other, and removes any latent sample bias due to different counts of left and right eyes in the dataset.


\subsection{B-scan classification with weak labels: RetiNet B}
\label{sec:blevel}

Recall that our data is inherently volumetric and that the amount of available data is relatively small. Given this challenging learning context, we will first learn features that are efficient at detecting typical 2D \ac{OCT} structures by learning to ``classify" B-scans. While our labels are only at the volume level, we propose to learn a B-scan level classifier by using approximately correct, or ``weak" labels to do so. In particular, we let $\hat{Y}^h_n = Y_n$, where $\hat{Y}^h_n$ is the weak label for $B^h_n$. Note that for all \textit{control} volumes, the $\hat{Y}^h_n=0$ labels indicate the lack of an \ac{AMD} diagnosis -- individual B-scan cross-sections should be pathology free. Conversely, volumes from subjects diagnosed with \ac{AMD} may contain a number of control or non-pathological B-scans. In particular, up to 50\% of the labels could be incorrect for such volumes.

To learn this classification function, we proceed by constructing a feed-forward \ac{CNN} whose architecture is illustrated in Fig.~\ref{fig:retinet}({\it top}). In this network, every single gray-scale $B_n^h$ is fed as input and passed through a set of 7 convolutional layers with small kernels ($3\times 3$, $5\times 5$) and max-pooling layers. We define this set of consecutive layers as the \textit{$FEATURE$} layers.

Following these layers, classification is achieved by using two consecutive fully connected layers and using a soft-max activation with two outputs for our two classes (\ie control and \ac{AMD}). We denote these latter layers as the \textit{$CLASSIFICATION$} layers. Throughout the entire network, convolutional and dense layers make use of leaky \acp{ReLU} activations. From this point on, we refer to this network as \BNET.

To train this network, we first begin by initializing all layer parameters randomly using Glorot Uniform sampling~\cite{Glorot2010}. We then make use of Extreme Learning ~\cite{Huang2006}, as it has been shown to increase regularization by forcing the convolutional layers to map to a broader features space. In practice, once the \textit{$CLASSIFICATION$} layers are initialized, we do not allow them to change. That is, we \textit{freeze} these layers and only allow the \textit{$FEATURE$} layers to be modified during the learning phase. In Sec.~\ref{sec:retinetCharact}, we show the effect of this learning strategy when compared to traditional regimes.


\subsection{Volume classification: RetiNet C}
\label{sec:clevel}

As we will show later in our experiments, the performance of the above network is limited, as it must learn from weak labels and does not make use of volumetric information to make a final decision. More so, it is not possible to test if the classification is in fact correct as the true label per B-scan is not known.

For this reason, we proceed to a second stage that attempts to classify the complete C-scan in one shot. Our proposed network is depicted in Fig.~\ref{fig:retinet}. Instead of setting each $B_n^h$ as an input channel, our network takes as input a vertical image of stacked B-scans, $M_n$, where
$$
M_n = \left[\begin{smallmatrix} B_n^1\\ \vdots \\ B_n^h \end{smallmatrix}\right],
$$
\noindent
resulting in a $\{w \times Dh \}$ sized image. Using the learned \textit{$FEATURE$} layers from the previous section, we include these into our new network as they are invariant to the size of the input and because ideally, these have learned what anatomical and pathological structures are relevant. These are then followed by 5 consecutive blocks of convolutional layers, average-pooling and batch normalization. Finally, we add the \textit{$CLASSIFICATION$} layer without transferring weights from \BNET. We define this network configuration as~\RTNET.

To train \RTNET, we freeze the \textit{$FEATURE$} layers, as these were learned on B-scans and should respond in the same way as above to preserve useful features extracted in the previous phase. The rest of the network is then trained using the true labels $Y_n$ to learn the remainder of the network layers.


\section{Evaluation}
\label{sec:experiments}

We now detail the performance of our strategy in the task of AMD classification in OCT volumes. We compare our approach to a number of existing state-of-the-art baselines coming from both the OCT pathology identification literature and the more general computer vision literature. We also provide qualitative results of our method, illustrating the different activation maps produced by our network and show how the different stages of our approach benefit the overall performance.

\subsection{Data set}
Our method was trained and evaluated on the publicly available dataset from Duke University~\cite{Farsiu2014}. This dataset was made available to find methods to define quantitative indicators for the presence of intermediate \ac{AMD}. In this set, 384 Spectral Domain OCT volumes are present, of which 269 volumes come from subjects with intermediate AMD while the remaining 115 subjects volumes were collected from healthy subjects. All scans are centered on the foveal pit. Each volume is acquired with 1000 A-scans per B-scan and 100 B-scans per volume. This results volume dimensions of \num{100 x 1000 x 500}$\mbox{px}^3$. In general, the volumes are not isotropic.

\subsection{Baselines}

To illustrate how each part of oue strategy influences overall performance, as well as to compare how our approach performs in contrast to other existing techniques in the literature, we now outline a number of baselines to which we will compare to directly:

\begin{itemize}
\item[-] \VGG: is the 19-layer variant of the deep CNN approach for image classification described in~\cite{Simonyan2015}. We pre-trained this network on the ImageNet dataset and fine-tuned the resulting filters using the \ac{OCT} dataset. To do that, we modified the receptive field of the network to match our B-scan resolution of 384x298 and exchanged the classification layer of the network with a fully-connected layer of size 2.
\item[-] \RESNET: similar to \VGG, we evaluated a pre-trained version of the 152-layer residual network described by He et al.~\cite{He2015}. Due to the highly tuned parameters of this network, we maintained the size of the receptive field at 224x224, opting instead to resize our input volume dimension to match. As before, we exchange the classification layer with a fully-connected layer of size 2.
\item[-] \DSNET: is a recent architecture network by Huang et al.~\cite{Huang2016}, which extends the residual network concept using a complete graph of skip connections. We implemented DenseNet with 3 dense blocks and a growth rate of 12, and trained the entire network on the \ac{OCT} dataset. 
\item[-] \VNET: is the patch-based classification scheme for pathological OCT identification described in Schlegl et al.~\cite{Schlegl2015}, which we re-implemented and trained on the \ac{OCT} dataset. Due to the lack of location information or per-voxel classifications in our ground truth data, we focused on the weak-labeling approach described in this same paper.
\end{itemize}

In addition to our complete \RTNET~approach, we also compare its performance to;
\begin{itemize}
\item[-] \BNETE: This consists of the RetiNet B classifier described in Sec.~\ref{sec:blevel} learned with Extreme Learning~\cite{Huang2006}. By comparing {\bf RetiNet} to this baseline, we can see the performance gain provided by the \RTNET~network construction.
\item[-] \BNET: Similarly to \BNETE, this classifier is identical in structure to the network of RetiNet B but trained without Extreme Learning.
\end{itemize}

In addition, we attempted to train both \VGG~and \RESNET~using only the OCT data, but given the large size of these networks and the small size of the dataset, this yielded in extremely poor classification methods. To avoid bias, we omit these methods from our experiments.

\subsection{Experimental setup}
We partition the dataset into five randomized, equi-sized subsets, using four for training and one for testing, for a total of five cross-validations per network. All methods were trained on the same partitions using the same folds. The random seed was preserved across all runs in order to remove any dataset-dependent bias.

We trained each network for a maximum of 100 epochs per fold, using early stopping with a patience of 15 epochs to avoid over-fitting~\cite{Prechelt1998,Larochelle2009}. We relied on the \textit{adadelta} algorithm~\cite{Zeiler2012} to optimize the parameters of each network. All networks except that of \VNET~were optimized by minimizing the categorical cross-entropy of their predictions versus the ground truth. \VNET~was optimized by minimizing the mean squared error, as described in~\cite{Schlegl2015}.

\VGG, \DSNET~and \RESNET~networks were trained and evaluated at the B-scan level using a weak labeling scheme, where the label of the complete C-scan was applied to each B-scan of that subject. The final C-scan classification prediction was defined as the mean score of the B-scan level predictions. The maximum achievable B-scan level accuracy is limited to roughly 94$pc$, due to mislabelings of individual B-scans (\ie the C-scan of an AMD patient may contain a number of healthy B-scans), as well as acquisition artifacts (\ie blinks).

\begin{figure}
    \centering
    \begin{subfigure}[b]{0.3\textheight}
        \includegraphics[width=\textwidth]{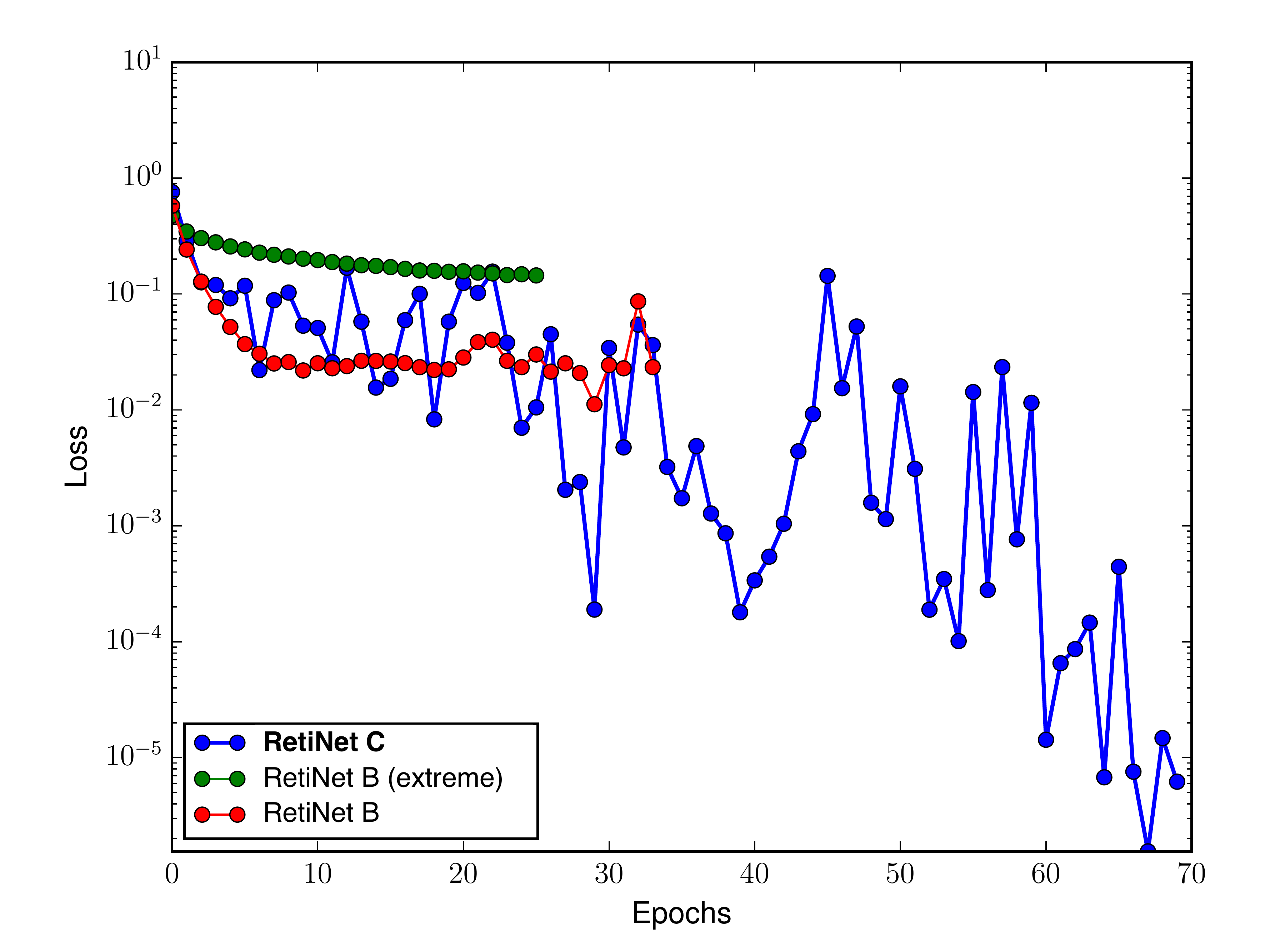}
        \caption{Training loss}
        \label{fig:learning_rates_train}
    \end{subfigure}
    \begin{subfigure}[b]{0.3\textheight}
        \includegraphics[width=\textwidth]{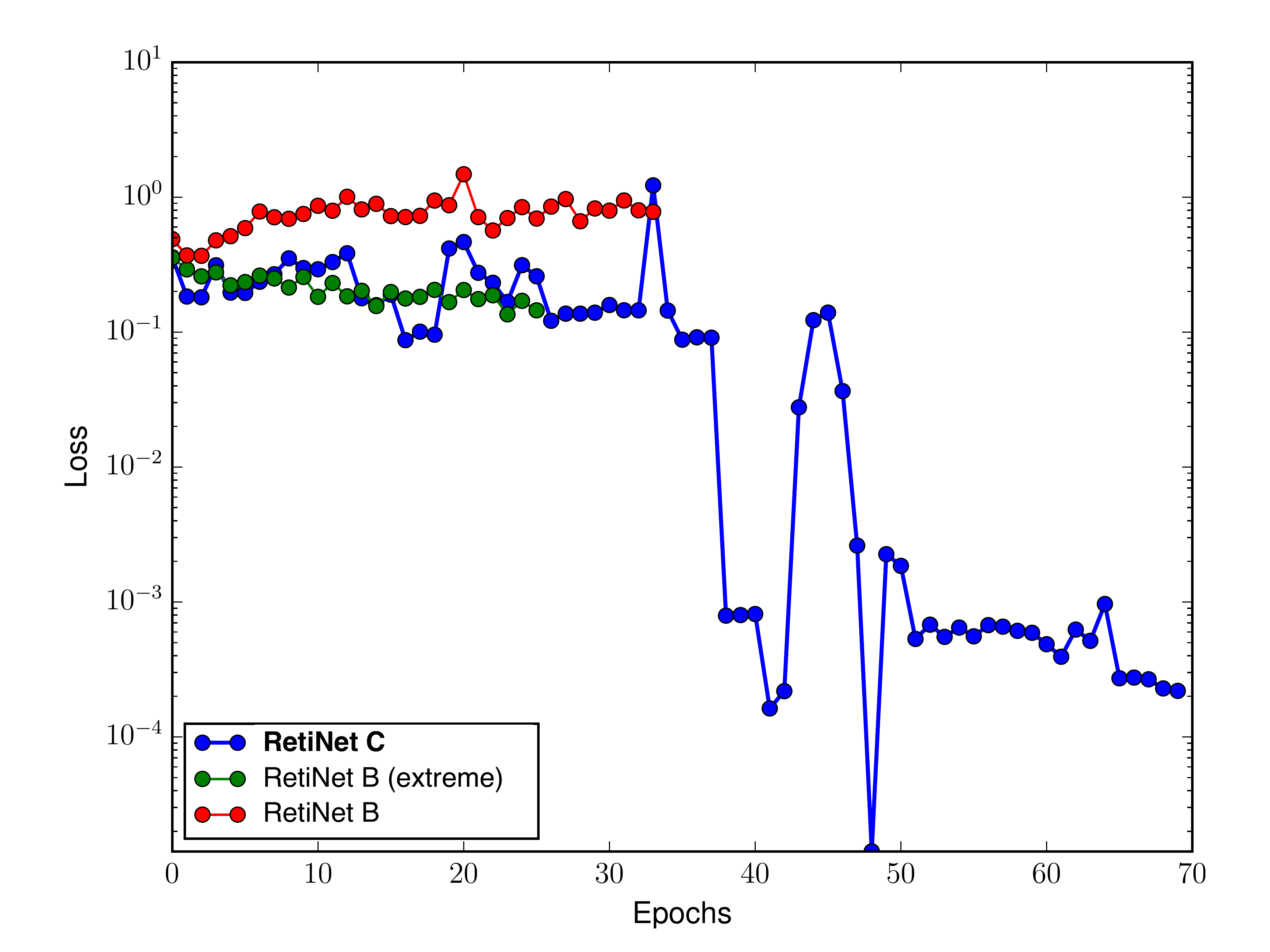}
        \caption{Validation loss}
        \label{fig:learning_rates_val}
    \end{subfigure}

    \caption{Comparison of training and validation loss for between {\bf RetiNet B}, {\bf RetiNet B Extreme} and {\bf RetiNet C}. Notice how the extreme learning configuration avoids overfitting compared to regular training. {\bf RetiNet C} is more effective than either with its loss decreasing throughout the training regime. The noisiness in {\bf RetiNet C} is most likely caused by the small amount of C-scans in the dataset, \ie 100x fewer compared to the amount of B-scans in the B configurations.}
    \label{fig:learning_rates}
\end{figure}

We provide a version of our \RTNET~implementation online\footnote{Visit \url{https://github.com/thefiddler/retinet} for an implementation of \RTNET.}. A complete list of the parameters used can be found in Table.~\ref{tab:parameters}. These were selected using experimental validation. In Fig.~\ref{fig:learning_rates}, we show the learning rate of our network with the above parameters on the training data and on the a validation set.

\begin{table}
	\centering
    \begin{tabular}{| l  c |}
    \hline
    Variable & Value \\ 
    \hline
    Cross-validation folds & 5 \\
    Training epochs & 100 \\ 
    Early stopping patience & 15\\
    Adadelta decay rate $\rho$ & 0.95\\
    Mini-batch size (\BNET/\RTNET) & 20 / 1 \\
    \hline
    \end{tabular}
    \caption{Summary of parameters used for training both \BNET~and \RTNET.}
    \label{tab:parameters}
\end{table}

\subsection{RetiNet characterization}
\label{sec:retinetCharact}

\begin{figure}
    \centering
    \begin{subfigure}[b]{0.45\textwidth}
        \includegraphics[width=\textwidth]{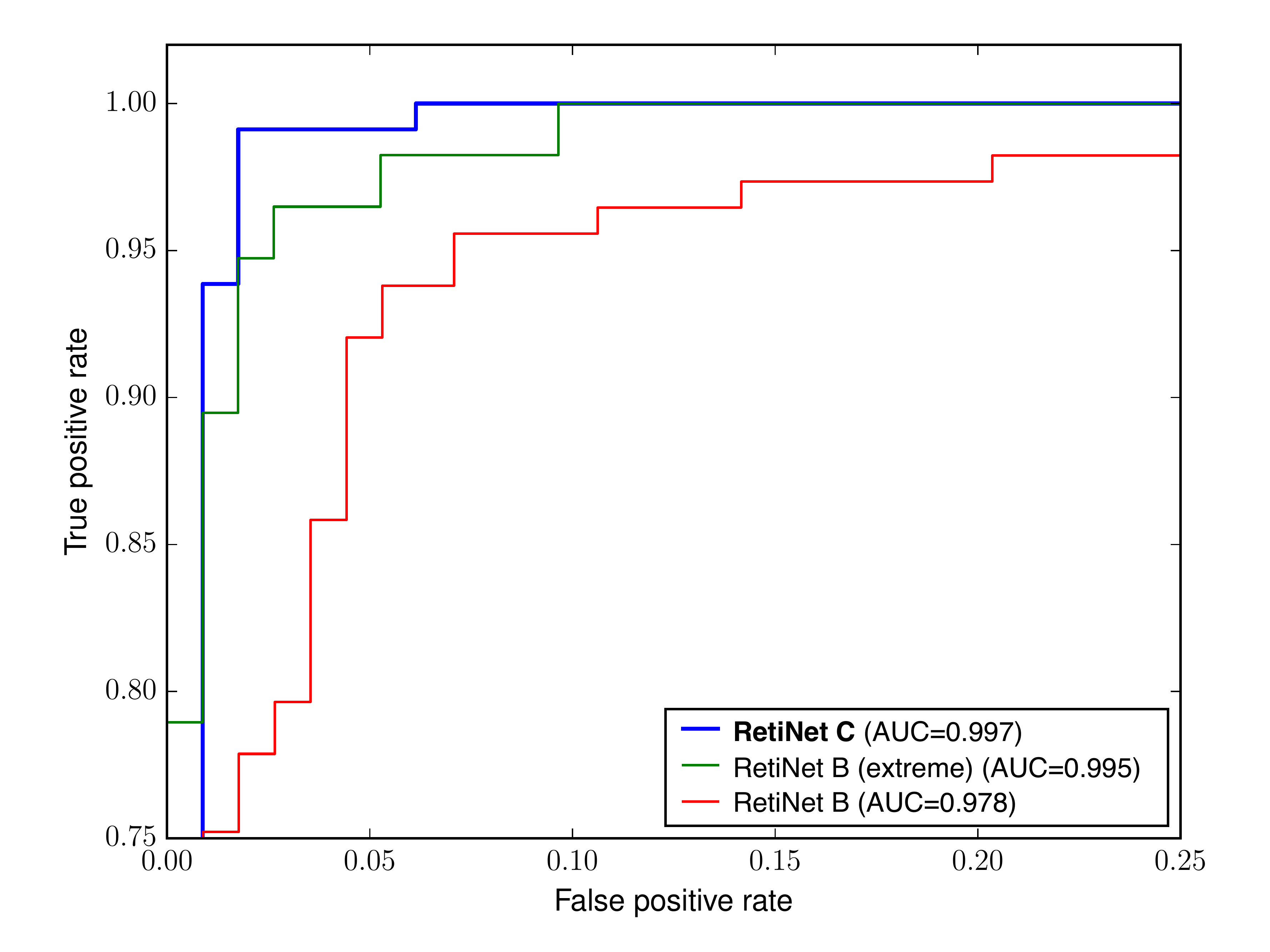}
        \caption{Receiver Operating Characteristic}
        \label{fig:retinet-cscan-roc}
    \end{subfigure}
	\begin{subfigure}[b]{0.45\textwidth}
        \includegraphics[width=\textwidth]{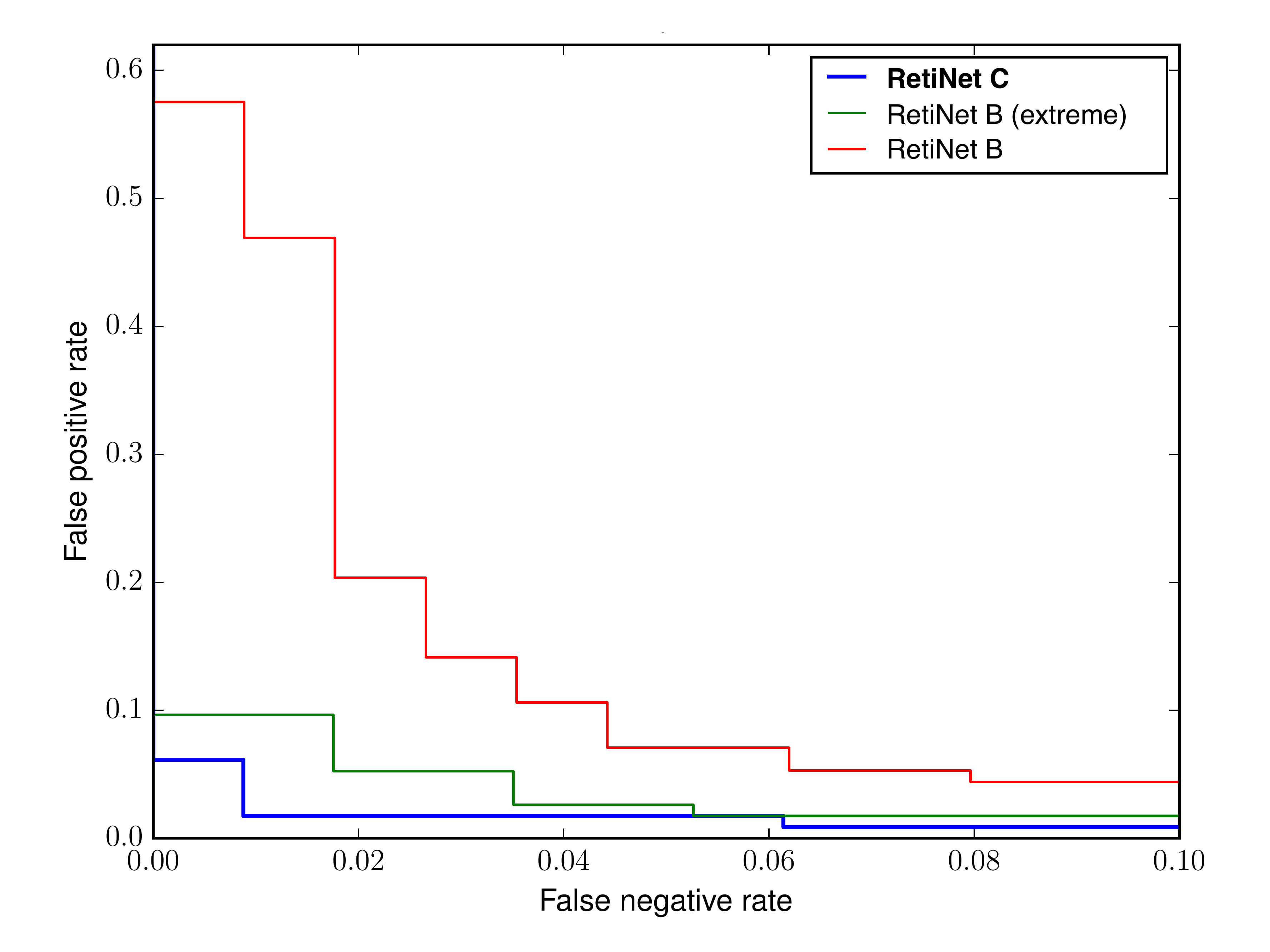}
        \caption{False Positive to False Negative ratio}
        \label{fig:retinet-cscan-fpfn}
    \end{subfigure}

    \caption{Comparison of \RTNET~variant to weakly-labeled~\BNETE~and~\BNET~configurations using ({\it top}) ROC curves and ({\it bottom}) False Negative Rate (FNR) versus False Positive Rate (FPR) curves. 
    }
\label{fig:retinet_charact}
\end{figure}

As an initial set of experiments, we are interested in characterizing the performance of our strategy. In Fig.~\ref{fig:retinet_charact}, we directly compare performances between \RTNET, \BNET~and \BNETE~ in terms of classification performance. The results shown in this figure are attained with 5-fold cross-validation.

First, we show in Fig.~\ref{fig:retinet_charact}({\it top}) the ROC curves of each of the three strategies. In addition to this traditional metric, we also show in Fig.~\ref{fig:retinet_charact}({\it bottom}) the False Negative Rate versus the False Positive Rate. This metric is more informative from a clinical perspective, as the clinical cost of classifying a pathological volume as a healthy volume is much higher than the other way around. In particular, we can see that for a 1\% false negative rate (\ie misclassifying pathological as healthy), \RTNET~has a 0.05\% false positive rate (\ie misclassifying healthy as pathological). This is interesting in the context of screening because it indicates that a human would need to evaluate pathological scans in any case and that the false positive rate indicates the reduced proportion of healthy scans one would still need to examine should one use an automatic classification algorithm. That is, allowing for a 1\% error in predicted subjects would only require a 5\% inspection of the healthy population. 

With this, we can observe that \BNET~has difficulty in learning correctly the volume labels given that it is trained on weak labels. As illustrated in the training and validation loss plots in Fig.~\ref{fig:learning_rates}, we can see that \BNET~effectively lacks generalization capabilities as it heavily overfits the data.

In contrast, \BNETE, with its Extreme Learning framework, allows for a stronger regularization and mitigates a significant amount of overfitting that is present with \BNET. As such, the difference in classification performance between \BNETE~and \RTNET~can be attributed to the weak labels and the 2D nature of the strategy.

\begin{figure*}[h]
\centering
\includegraphics[width=0.99\textwidth]{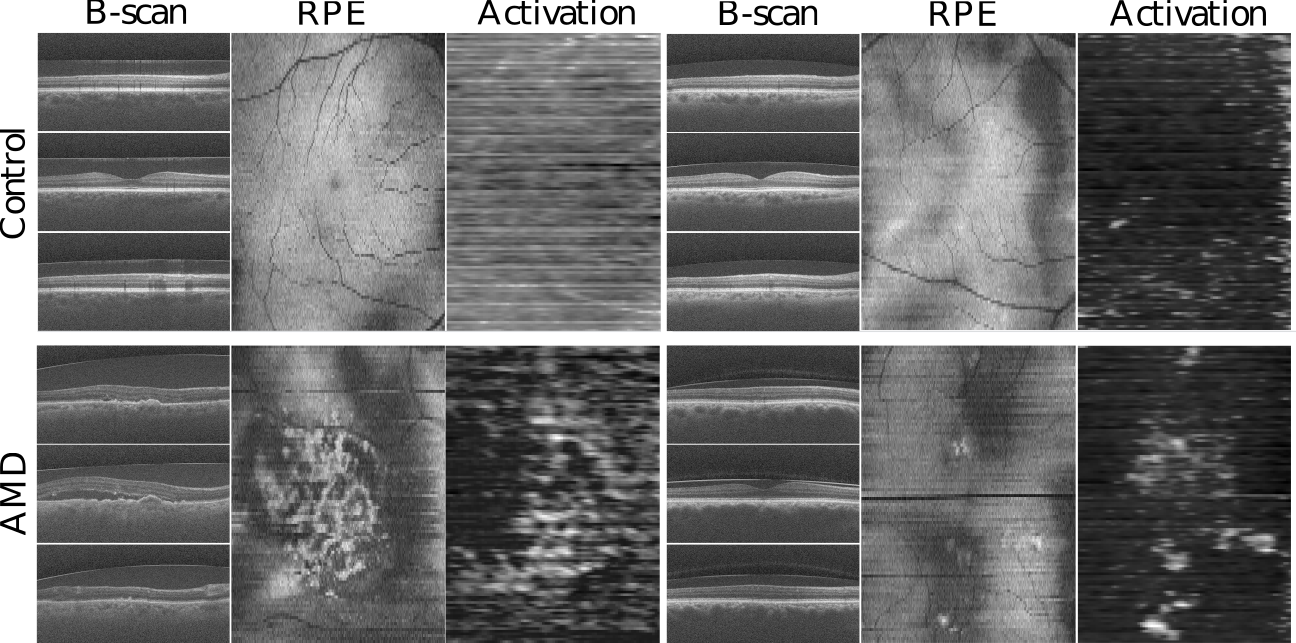}
\caption{Activation maps of the last convolutional \RTNET ~layer for four different volumes. The \textit{top} row depicts two control subjects, while the \textit{bottom} row depicts two \ac{AMD} subjects. Each subject is described in three columns: \textit{(left)} Three B-scans at slice 25, 50 and 75; \textit{(center)} the top-down Fundus reconstruction of the \ac{RPE} layer; \textit{(right)} Activation map of the last convolutional layer of the \RTNET~configuration. The activation maps highlights pathological structures relevant for \ac{AMD}.}
\label{fig:activation_maps}
\end{figure*}

In this sense, these results highlight that \RTNET~overcomes the lack of B-scan level labels and that weak labels can be exploited at the C-scan level. To illustrate what our network learns however, we can visualize the network activation maps in Fig.~\ref{fig:activation_maps}. Here we show four (two healthy and two AMD) examples of volumes and how our network responds to them. For each case, we show three B-scans from a volume ($h=25,50,75$), the associated fundus view of the RPE retinal layer and the projection of the activation map of the last convolutional layer \RTNET. In particular, we can see for the AMD cases that the activation maps are responding very strongly at different locations of the volume and differently to healthy volumes. 

Last, the learning of this two-stage network appears to learn even after 70 epochs, indicating that the overfitting is most likely limited. At the same time, we notice that learning is in general consistently noisy with our framework and this is most likely due to the limited number of C-scans available during training. 

\begin{figure*}
\centering
\includegraphics[trim={0 15cm 0 0},clip,width=0.95\textwidth]{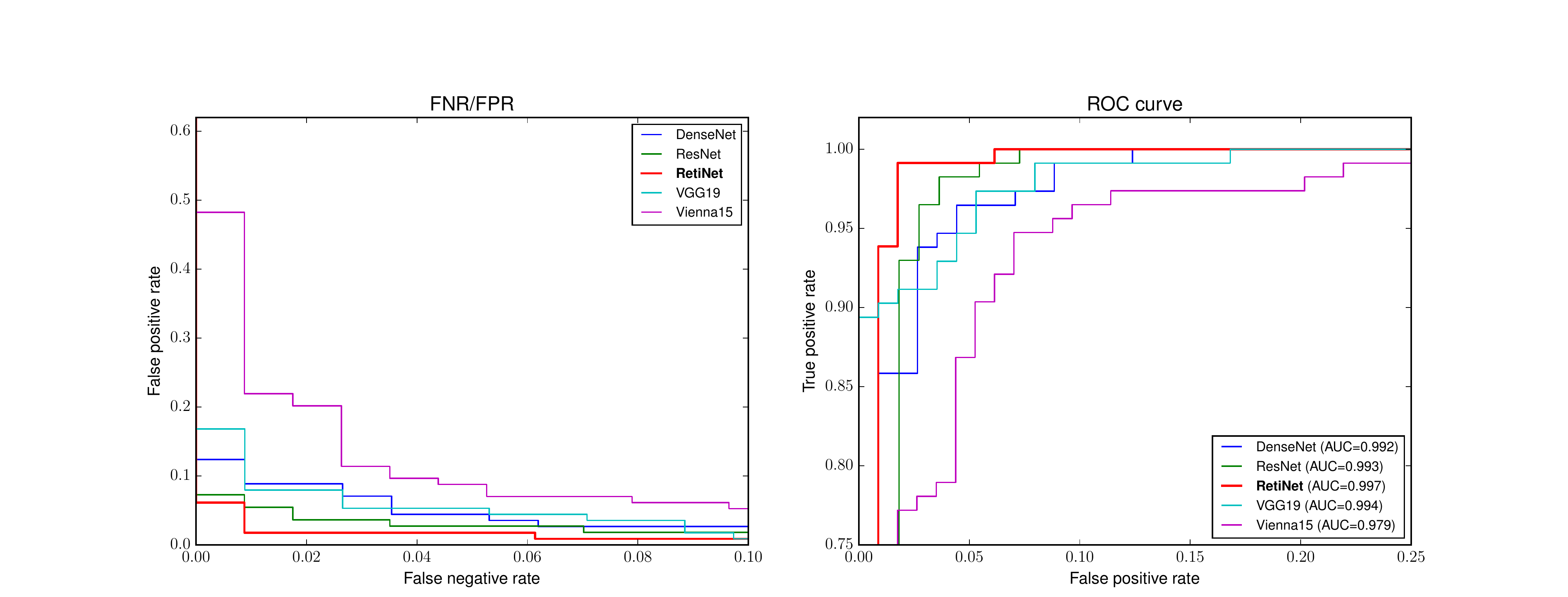}
\caption{Comparison of \RTNET~mosaic to state-of-the-art image classification networks: \VGG, \RESNET~and \DSNET, as well as the weakly-labeled approach described in Schlegl et al.~\cite{Schlegl2015}. \VGG~and \RESNET~are pre-trained on ImageNet and fine-tuned on the OCT dataset. The results are after a 5-fold cross-validation. Comparisons are with respect to ({\it left}) False Negative Rate (FNR) versus False Positive Rate (FPR) curves and ({\it right}) ROC curves.}
\label{fig:baselineComp}
\end{figure*}


\subsection{Baseline comparison}

Fig.~\ref{fig:baselineComp} outlines the performance of \RTNET~and the baseline methods in terms of ROC, as well as FNR/FPR. Across both metrics, \RTNET~appears to outperform these baselines.

A number of interesting conclusions can in addition be drawn from these results. First, off-the-shelf computer vision networks that perform exceptionally well on natural images \VGG,~\DSNET~and \RESNET, have a strong tendency to overfit this data. Second, \DSNET~performs very similarly to both \VGG~and \RESNET~even though it is trained from scratch and converges very quickly. Both \VGG~and \RESNET~could not be successfully trained from scratch however due to the relatively small dataset. Last, the \VNET~approach, which was specifically developed for this application, appears to have difficulties generalizing when training with weak labels. This is consistent with authors conclusion as well~\cite{Schlegl2015}.

From our experiments, we can see that \RTNET~achieves an \ac{AUC} of of 99.7\%. This compares favorably to the semi-automatic method of Farsiu et al.~\cite{Farsiu2014}, which achieved an \ac{AUC} of 99.17\% in their reported best case and the automatic method of Venhuizen et al.~\cite{Venhuizen2015} which achieved an \ac{AUC} of 98.4\%. While we do not compare to these methods directly, we report these scores from their published results on the same data.

In this sense, it appears as though \RTNET~provides a stable strategy capable of leveraging volumetric information and only uses the volume level labels for training. Fig.~\ref{fig:examples} shows a few examples where \RTNET~correctly and incorrectly predicts different volume labels. 

\section{Conclusions}
\label{sec:conc}

In this article, we have proposed a novel strategy for automatic identification of \ac{AMD} in OCT volumes. Our strategy is advantageous as it only requires volume-level labels as opposed to cross-sectional labels, making it far easier to train from a groundtruth acquisition point of view. Our approach involves a novel two-stage deep learning architecture that in the first phase, focuses on learning features that are domain specific and then focuses on the volume classification task in the latter phase.

We validated our approach using publicly available OCT data and compared the performance of our method against both techniques from the OCT domain and the computer vision literature. We showed that not only does our approach do well in terms of ROC performance, but that it also does well with respect to a more clinically relevant metric.

\begin{figure*}
    \centering
    \begin{subfigure}[b]{0.45\textwidth}
        \includegraphics[width=\textwidth]{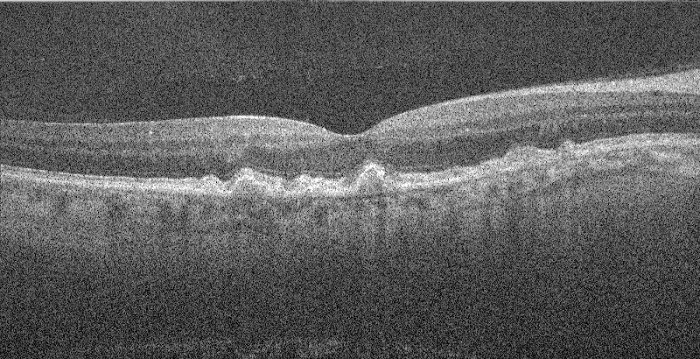}
        \caption{True positive}
        \label{fig:TP}
    \end{subfigure}
    \begin{subfigure}[b]{0.45\textwidth}
        \includegraphics[width=\textwidth]{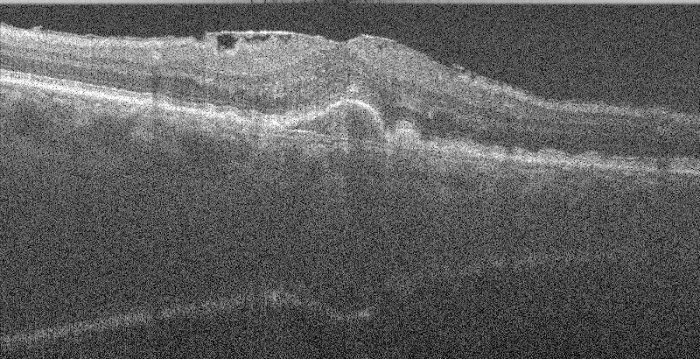}
        \caption{True positive}
        \label{fig:TP2}
    \end{subfigure}

    \bigskip

    \begin{subfigure}[b]{0.45\textwidth}
        \includegraphics[width=\textwidth]{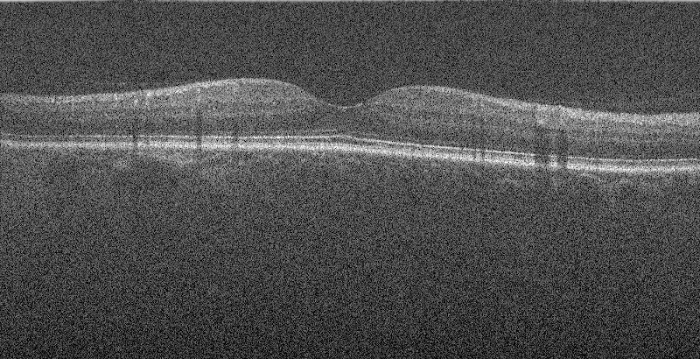}
        \caption{True negative}
        \label{fig:TN}
    \end{subfigure}
    \begin{subfigure}[b]{0.45\textwidth}
        \includegraphics[width=\textwidth]{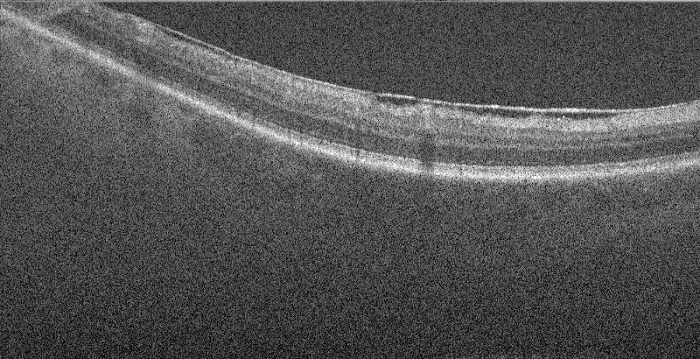}
        \caption{True negative}
        \label{fig:TN2}
    \end{subfigure}

    \bigskip

    \begin{subfigure}[b]{0.45\textwidth}
        \includegraphics[width=\textwidth]{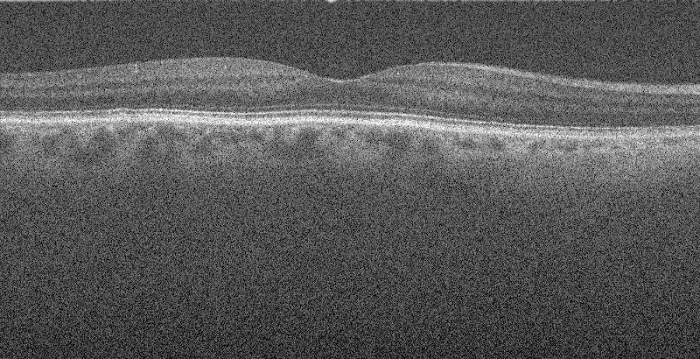}
        \caption{False positive}
        \label{fig:FP}
    \end{subfigure}
    \begin{subfigure}[b]{0.45\textwidth}
        \includegraphics[width=\textwidth]{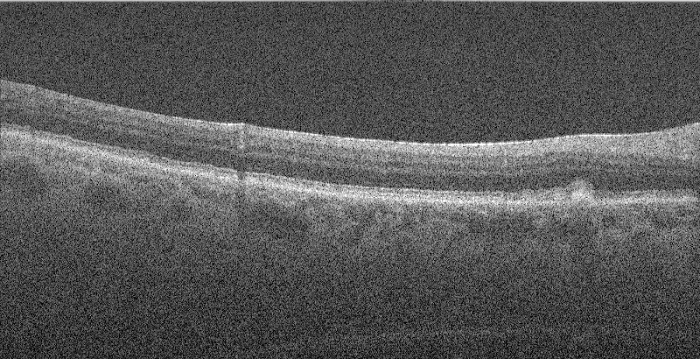}
        \caption{False negative}
        \label{fig:FN}
    \end{subfigure}
    
    \caption{Example B-scans from correctly and incorrectly classified volumes. While (a-d) show correctly identified cases, (e-f) are incorrectly classified. Surprisingly, our approach correctly identifies (d) as non-AMD, even though it illustrates an epiretinal membrane and vitreoretinal traction, neither of which is AMD.}
    \label{fig:examples}
\end{figure*}

This being said, our method still has difficulties identifying mild AMD cases as shown in Fig.~\ref{fig:examples}(e) where the difference between healthy and pathological is visibly challenging. In this sense, we will focus in the future on developing strategies for identifying early-stages of the disease and we will look at how diseases differ to one another. 

\bibliographystyle{IEEEtran}
\bibliography{bibliography.bib}

\end{document}